\documentclass[letterpaper, 10 pt, conference]{ieeeconf}  

\IEEEoverridecommandlockouts                              

\usepackage{graphics} 
\usepackage{epsfig} 
\usepackage{times} 
\usepackage{amsmath} 
\usepackage{amssymb}  
\usepackage{cite}
\usepackage{bm}
\usepackage{url}
\usepackage{float}
\usepackage{xcolor} 
\usepackage[symbol]{footmisc}

\title{\LARGE \bf
Combining Movement Primitives with Contraction Theory
}
 
\author{Moses C. Nah,$^{1}$ Johannes Lachner,$^{1,2}$ Neville Hogan,$^{1,2}$ and Jean-Jacques Slotine$^{1,2}$ 
\thanks{$^{1}$Department of Mechanical Engineering, Massachusetts Institute of Technology, Cambridge, MA, USA.}
\thanks{$^{2}$Department of Brain and Cognitive Sciences, Massachusetts Institute of Technology, Cambridge, MA, USA.}
\thanks{ MCN was supported in part by a Mathworks Fellowship. JL was supported in part by the MIT-Novo Nordisk Artificial Intelligence Postdoctoral Fellows Program. } 
\thanks{This work has been submitted to the IEEE for possible publication. Copyright may be transferred without notice, after which this version may no longer be accessible.}
}

\begin{document}

\newcommand{\revisionone}[1]{\textcolor{black}{#1}}

\maketitle
\thispagestyle{empty}
\pagestyle{empty}

\begin{abstract}
This paper presents a modular framework for motion planning using movement primitives.
Central to the approach is Contraction Theory, a modular stability tool for nonlinear dynamical systems.
The approach extends prior methods by achieving parallel and sequential combinations of both discrete and rhythmic movements, while enabling independent modulation of each movement.
This modular framework enables a divide-and-conquer strategy to simplify the programming of complex robot motion planning.
Simulation examples illustrate the flexibility and versatility of the framework, highlighting its potential to address diverse challenges in robot motion planning.
\end{abstract}

\section{INTRODUCTION}
Motion planning using movement primitives has been a popular approach in robotics \cite{williamson1999robot, ijspeert2002learning,schaal2007dynamics,righetti2008pattern,pastor2009learning,hogan2012dynamic,ijspeert2013dynamical,koutras2020correct,khadivar2021learning,figueroa2022locally,billard2022learning, saveriano2023dynamic,nah2024robot,abu2024unified}. 
Often referred to as a ``dynamical systems approach'' \cite{billard2022learning}, it encodes trajectories (or ``movement primitives'') as dynamical systems with specific attractor dynamics. 
For instance, a dynamical system with a stable fixed point attractor generates discrete point-to-point movement \cite{khansari2011learning,koutras2020novel}, while one with a stable limit cycle is used to generate rhythmic, repetitive actions \cite{williamson1999robot,abu2024learning}.

Using dynamical systems for motion planning offers several advantages.
The stability  arising from the attractor dynamics provides robustness against errors and external disturbances \cite{ijspeert2013dynamical,saveriano2023dynamic}. 
Fast online reactive control can be achieved in dynamically changing environments \cite{billard2022learning}, e.g., real-time obstacle avoidance \cite{hoffmann2009biologically,khansari2012dynamical}.
The approach also facilitates robot learning.
Methods such as Learning from Demonstration (LfD) \cite{billard2022learning} and Imitation Learning \cite{ijspeert2013dynamical,saveriano2023dynamic,nah2024robot} allow robots to learn a desired attractor dynamics from few demonstrations.\footnote{Although both methods are based on the same dynamical systems approach, they exhibit clear technical differences. For further details, readers may refer to Section 3.5 of \cite{saveriano2023dynamic}.} 

These favorable properties of using dynamical systems enabled success in a variety of applications \cite{peters2008natural,kober2009learning,hoffmann2009biologically,khansari2012dynamical,zhou2017task,billard2019trends,huber2022fast}. 
Nevertheless, a key limitation is the restricted range of movements that can be generated. 
Research has been dominated by generating a (sequence of) discrete point-to-point movement(s) \cite{billard2022learning}.
A method to combine both discrete and rhythmic movement primitives has not been thoroughly addressed \cite{saveriano2023dynamic,nah2024robot}, despite its practical value for robot control (e.g., polishing or sanding a surface \cite{khadivar2021learning} or peg-in-hole assembly \cite{lachner2024divide}). 
A fundamental reason is because the stability proof of prior approaches is based on (strictly) stable Lyapunov functions \cite{khansari2011learning,figueroa2022locally}, which precludes limit cycle attractors \cite{strogatz2018nonlinear}. 
From Strogatz \cite{strogatz2018nonlinear}, ``\textit{If a (strictly stable) Lyapunov function exists, then closed orbits are forbidden.}''
In fact, this issue also exists for other Lyapunov-based motion planning methods \cite{burridge1999sequential,majumdar2017funnel}. 
While ways to combine both discrete and rhythmic movements have been proposed \cite{degallier2006movement,ernesti2012encoding,khadivar2021learning}, they do not allow independent modification of each movement, since the combined movement is learned using a single dynamical system.
\revisionone{This limitation reduces flexibility in motion planning, necessitating movements to be learned from scratch instead of being constructed by combining existing ones. 
One might circumvent this issue by separately constructing dynamical systems for discrete and rhythmic movements, integrating their results separately, and then combining the outcomes.
However, this approach renounces the advantages of using dynamical system, such as robustness against errors and disturbances, adaptability and reactiveness against dynamically changing environments. }

In this paper, we extend Dynamic Movement Primitives \cite{ijspeert2013dynamical,saveriano2023dynamic} by proposing a method to combine both discrete and rhythmic movement primitives.
The key is to use Contraction Theory \cite{lohmiller1998contraction,slotine2003modular,perk2006motion,wensing2017sparse}, a modular stability tool to analyze nonlinear dynamical systems. 
We show that both discrete and rhythmic movements can be combined while preserving independent modification of each movement. 
Stability is proven by Contraction Theory. Using the fact that the condition of contraction is preserved (or closed) under various system combinations, we show that a combination of discrete and rhythmic movements is simply a combination of contracting and transverse-contracting systems, respectively. 
\revisionone{With stability analysis provided by Contraction Theory, the advantages of using dynamical systems are preserved.}

The contributions of this paper are listed below:
\begin{itemize}
    \item A definition of \revisionone{movement} primitives for both discrete and rhythmic movements.
    \item A method for parallel and sequential combinations of \textit{both} discrete and rhythmic movements, while allowing spatial and temporal scaling of each movement. 
    \item A stability proof of the presented approach using Contraction Theory. 
\end{itemize}
Simulation examples are presented with possible extensions for actual robot control.\footnote{The codes are available in Github: \url{https://github.com/mosesnah-shared/modularDMP}}

\section{Contraction Theory}
In this Section, an overview of Contraction Theory is provided. 
The key theorems are presented; their proofs are omitted. 
For further details, see \cite{lohmiller1998contraction,slotine2001modularity,slotine2003modular,tsukamoto2021contraction,bullo2022contraction}. 

Throughout this paper, an $n$-dimensional nonlinear dynamical system is considered:
\begin{equation}\label{eq:nonlinear_dynamical_system}
    \dot{\mathbf{x}}(t) = \mathbf{f}(\mathbf{x}(t), t)
\end{equation}
In this equation, $t\in\mathbb{R}_{\ge0}$ is time; $\mathbf{x}(t)\equiv \mathbf{x}\in\mathbb{R}^{n}$ is the state (trajectory) of the system; $\mathbf{f}:\mathbb{R}^{n}\times \mathbb{R}\rightarrow \mathbb{R}^n$ is a vector field. 
All maps and quantities are assumed to be smooth unless specified. 
To avoid clutter, the time argument is often omitted. 
 
Contraction Theory, introduced by Lohmiller and Slotine \cite{lohmiller1998contraction}, considers whether trajectories of a nonlinear dynamical system converge toward a single, nominal trajectory. 
A system that exhibits such convergent behavior is referred to as contracting.
For contracting systems, initial conditions or temporary disturbances are ``forgotten'' (exponentially fast), and the asymptotic behavior of the dynamical system converges to a nominal behavior. 
In contrast to Lyapunov Stability Theory, a key aspect of Contraction Theory is that \textit{a prior} knowledge about the nominal trajectory is not required. 
This simplifies  stability proofs for nonlinear dynamical systems, where an analytical solution of the nominal trajectory may be difficult to derive \cite{tsukamoto2021contraction,bullo2022contraction}.

\subsection{Definitions}
In this Section, definitions of contracting and transverse contracting systems are provided. 
\subsubsection{Contraction}
A nonlinear dynamical system is contracting in region $\mathcal{C}\subseteq \mathbb{R}^{n}$, if $\forall t \in \mathbb{R}_{\ge 0}$ and $\forall \mathbf{x}\in \mathcal{C}$, there exists both a symmetric positive definite matrix $\mathbf{M}(\mathbf{x}, t)\equiv \mathbf{M} \in\mathbb{R}^{n\times n}$ and a positive constant $\lambda\in\mathbb{R}_{>0}$ that satisfy:
\begin{equation}\label{eq:contracting_condition}
    \dot{\mathbf{M}} + \Big(\frac{\partial \mathbf{f}}{\partial \mathbf{x}}\Big)^{\top}\mathbf{M} + \mathbf{M}\frac{\partial \mathbf{f}}{\partial \mathbf{x}} \preceq -2\lambda \mathbf{M}
\end{equation}
where $\frac{\partial \mathbf{f}}{\partial \mathbf{x}}\equiv \frac{\partial \mathbf{f}}{\partial \mathbf{x}}(\mathbf{x},t)\in\mathbb{R}^{n\times n}$ is a Jacobian matrix of $\mathbf{f}$.
Region $\mathcal{C}$, matrix $\mathbf{M}$, and positive constant $\lambda$ are referred to as the ``contraction region,'' ``contraction metric,'' and ``contraction rate'' of the system, respectively. 

\subsubsection{Transverse Contraction}
For autonomous dynamical systems with stable limit cycles, the system is not contracting along the phase variable \cite{wensing2017sparse}.
In this case one can analyze contraction along a direction orthogonal (with respect to a contraction metric) to the flow $\mathbf{f}(\mathbf{x})$.

Consider an autonomous dynamical system $\dot{\mathbf{x}}=\mathbf{f}(\mathbf{x})$. 
The system is said to be transverse contracting in region $\mathcal{C}_{\mathcal{T}} \subseteq \mathbb{R}^{n}$, if $\forall \mathbf{x}\in\mathcal{C}_{\mathcal{T}}$, there both exist a time-invariant symmetric positive definite matrix $\mathbf{M}_{\mathcal{T}}(\mathbf{x})\equiv \mathbf{M}_\mathcal{T}\in\mathbb{R}^{n\times n}$ and a positive constant $\lambda_{\mathcal{T}} \in \mathbb{R}_{>0}$ that satisfy:
\begin{equation}\label{eq:transverse_contracting}
    \delta \mathbf{x}^{\top} \Big\{ \dot{\mathbf{M}}_\mathcal{T} + \Big(\frac{\partial \mathbf{f}}{\partial \mathbf{x}}\Big)^{\top}\mathbf{M}_\mathcal{T} + \mathbf{M}_\mathcal{T}\frac{\partial \mathbf{f}}{\partial \mathbf{x}} + 2\lambda_\mathcal{T}  \mathbf{M}_\mathcal{T} \Big\} \delta \mathbf{x} \le 0
\end{equation}
for all $\delta \mathbf{x}\in\mathbb{R}^{n}\setminus \{ \mathbf{0} \}$ satisfying the orthogonality condition $\delta\mathbf{x}^{\top}\mathbf{M}_\mathcal{T}(\mathbf{x})\mathbf{f}(\mathbf{x})$. Region $\mathcal{C}_{\mathcal{T}}$, matrix $\mathbf{M}_{\mathcal{T}}$, and positive constant $\lambda_{\mathcal{T}}$ are referred to as the ``transverse contraction region,'' ``transverse contraction metric,'' and ``transverse contraction rate'' of the system, respectively. 

\subsection{Modular Properties of Contracting Systems}
Contraction (or transverse contraction) are automatically preserved through various system combinations \cite{lohmiller1998contraction,slotine2001modularity,slotine2003modular}. 
All results are global unless otherwise specified.

\subsubsection{Parallel Combination \cite{lohmiller1998contraction}}\label{subsubsec:parallel_combination}
Consider two dynamical systems with identical dimensions, that are contracting \textit{under the same contraction metric} $\mathbf{M}(\mathbf{x})$:
\begin{equation*}
    \dot{\mathbf{x}}(t) = \mathbf{f}_1(\mathbf{x},t), ~~~~~ \dot{\mathbf{x}}(t) = \mathbf{f}_2(\mathbf{x},t)
\end{equation*}
Then $\forall t\in\mathbb{R}_{\ge 0}$, a parallel combination of these two contracting systems also contracts under the same contraction metric $\mathbf{M}(\mathbf{x})$:
\begin{equation}
    \alpha_1(t), \alpha_2(t) \in\mathbb{R}_{>0}: \dot{\mathbf{x}} = \alpha_1(t)\mathbf{f}_1 (\mathbf{x}, t )+ \alpha_2(t)\mathbf{f}_2(\mathbf{x}, t )
\end{equation}
The contraction rate of the combined dynamical system is $\alpha_1(t)\lambda_1 + \alpha_2(t)\lambda_2$, where $\lambda_1$ and $\lambda_2$ are the contraction rates of $\mathbf{f}_1$ and $\mathbf{f}_2$, respectively.

\subsubsection*{(Remark 1)}
As stated by Slotine \cite{slotine2003modular}, one example is a parallel combination of movement primitives. In detail, consider a movement primitive input $\mathbf{p}_i(\mathbf{x},t)\in\mathbb{R}^{p}$, where for each $i\in\{1,2,\cdots, N_p\}$, the following dynamical system contracts under the same contraction metric $\mathbf{M}(\mathbf{x})$:
\begin{equation}
    \dot{\mathbf{x}}(t) = \mathbf{f}(\mathbf{x},t) + \mathbf{B}(\mathbf{x},t)\mathbf{p}_{i}(\mathbf{x},t)
\end{equation}
In this equation, $\mathbf{B}(\mathbf{x},t)\in\mathbb{R}^{n\times p}$ is the input weighting matrix. Then, a combination of these $N_p$ primitives:
\begin{gather}
    \forall i \in \{1,2,\cdots, N_p\} ~~~~~ \alpha_i(t)\ge 0 ~~~~~ \sum_{i=1}^{N_p}\alpha_i(t)=1 \nonumber \\
    \dot{\mathbf{x}}(t) = \mathbf{f}(\mathbf{x},t) + \sum_{i=1}^{N_p}\mathbf{B}(\mathbf{x},t)\Big( \alpha_i(t)\mathbf{p}_i(\mathbf{x},t)\Big) 
\end{gather}
also contracts under the same contraction metric $\mathbf{M}(\mathbf{x})$ with contraction rate $\sum_{i=1}^{N_p}\alpha_i(t)\lambda_i$, where $\lambda_i$ is the contraction rate of the $i^{\text{th}}$ primitive input. 

Moreover, given a duration $T\in\mathbb{R}_{>0}$, if a $T$-periodic primitive input---potentially derived from the phase output of a transverse contracting system---is included, $\mathbf{x}(t)$ converges to a \textit{unique} $T$-periodic solution \cite{bullo2022contraction}.

\subsubsection{Hierarchical Combination \cite{lohmiller1998contraction,simpson2014contraction}}
Consider two dynamical systems, of possibly different dimensions and contraction metrics, and their hierarchical combination, i.e., the system dynamics  $\dot{\mathbf{x}}_1(t)=\mathbf{f}_1(\mathbf{x}_1, t)$ and $\dot{\mathbf{x}}_2(t)=\mathbf{f}_2(\mathbf{x}_1, \mathbf{x}_2, t)$, leading to virtual dynamics of the form:
\begin{equation}
    \begin{bmatrix}
        \delta \dot{\mathbf{x}}_1(t) \\
        \delta \dot{\mathbf{x}}_2(t) \\
    \end{bmatrix} 
    = 
    \begin{bmatrix}
        \mathbf{F}_{11}(\mathbf{x}_1)  & \mathbf{0} \\
        \mathbf{F}_{12}(\mathbf{x}_1,\mathbf{x}_2) & \mathbf{F}_{22}(\mathbf{x}_1, \mathbf{x}_2)  \\
    \end{bmatrix}     
    \begin{bmatrix}
        \delta \mathbf{x}_1(t) \\
        \delta \mathbf{x}_2(t) \\
    \end{bmatrix}     
\end{equation}
If $\mathbf{f}_1$ is contracting, $\mathbf{f}_2$ is contracting (for each fixed $\mathbf{x}_1$), and $\mathbf{F}_{12}$ is bounded, then the overall system is contracting, i.e., $\delta\mathbf{x}_1(t), \delta\mathbf{x}_{2}(t)\rightarrow \mathbf{0}$.
The contraction rate of the overall system is the slowest of the individual contraction rates.

\subsubsection*{(Remark 2)}
If either of the vector fields $\mathbf{f}_1$ or $\mathbf{f}_2$ is transverse contracting (possibly with different transverse contraction metrics), then the overall system is transverse contracting  \cite{manchester2014transverse}. 

\subsubsection{Sequential Combination \cite{wang2005partial}}\label{subsubsec:sequential_combination}
Consider a dynamical system $\dot{\mathbf{x}}_1(t)=\mathbf{f}(\mathbf{x}_1, t)$ which is one-way coupled to another dynamical system with identical dimension via input $\mathbf{u}$:
\begin{align}
    \begin{split}
        \dot{\mathbf{x}}_1(t) &= \mathbf{f}(\mathbf{x}_1, t) \\
        \dot{\mathbf{x}}_2(t) &= \mathbf{f}(\mathbf{x}_2, t) + \mathbf{u}(\mathbf{x}_1) - \mathbf{u}(\mathbf{x}_2)
    \end{split}
\end{align}
If vector field $\mathbf{f}-\mathbf{u}$ is contracting, then $\mathbf{x}_2(t)\rightarrow \mathbf{x}_1(t)$.

\section{Dynamic Movement Primitives}
This Section presents an overview of Dynamic Movement Primitives (DMP). For further details see \cite{ijspeert2013dynamical,saveriano2023dynamic,nah2024robot}.

DMP, introduced by Ijspeert and Schaal \cite{ijspeert2002movement,schaal2003control}, generates trajectories of arbitrary complexity by using dynamical systems with specific attractors. 
DMP consists of three components: a canonical system, a nonlinear forcing term and a transformation system. 
Each component plays a distinct role in trajectory generation, as will be detailed in this Section.

Several types of DMP exist, depending on the space in which the trajectory resides. 
For instance, to generate trajectories for spatial orientation (i.e., on the $\text{SO}(3)$ manifold), DMP which accounts for the manifold structure of $\text{SO}(3)$ should be used \cite{koutras2020correct,abu2024unified}. 
While the approach we present can be generalized to these cases, in this paper, we focus on DMP which generate trajectories on $\mathbb{R}^{n}$.

\subsection{Canonical System}\label{subsec:canonical_system}
A canonical system serves as an ``internal clock'' of movement primitives. 
It provides a notion of phase without an explicit time parameterization \cite{ijspeert2013dynamical}.
Different types of canonical systems are used for discrete and rhythmic movements, as discrete movement requires a temporal phase, unlike the periodic phase used in rhythmic movement.

\subsubsection{For Discrete Movement}
A canonical system for discrete movement, $s_{d}:\mathbb{R}_{\ge 0} \rightarrow \mathbb{R}_{\ge 0}$ is defined by a first-order stable linear system:
\begin{equation}\label{eq:canonical_system_discrete}
    \tau_{d}\dot{s}_{d}(t) = -\alpha_{s} s_{d}(t)
\end{equation}
In this equation, $\alpha_s, \tau_{d}\in\mathbb{R}_{>0}$ determine the time constant; the analytical solution is given by $s_d(t)=\exp(-\frac{\alpha_s}{\tau_d}t) s_d(0)$; while any initial conditions $s_d(0)$ can be used, the value is typically chosen to be $s_d(0)=1$ \cite{saveriano2023dynamic}.

\subsubsection{For Rhythmic Movement}
A canonical system for rhythmic movement, $s_{r}:\mathbb{R}_{\ge 0}\rightarrow [0, 2\pi)$ is a periodic function, defined by the phase variable of the Andronov-Hopf oscillator\revisionone{---a nonlinear dynamical system with a stable limit cycle attractor}:
\begin{equation}\label{eq:canonical_system_andronov}
    \begin{bmatrix}
        \dot{x}_1 \\
        \dot{x}_2 
    \end{bmatrix}
    = 
    \begin{bmatrix}
        \gamma^2 - (x_1^2 + x_2^2) & -1/\tau_{r} \\
        1/\tau_{r} & \gamma^2 - (x_1^2 + x_2^2)
    \end{bmatrix}    
    \begin{bmatrix}
        x_1 \\
        x_2 
    \end{bmatrix}
\end{equation}
In this equation, $\gamma, \tau_r \in\mathbb{R}_{>0}$; $x_1(t), x_2(t)\in\mathbb{R}$ are the state-variables. For non-zero initial conditions, all trajectories converge to a stable limit cycle with radius $\gamma$ rotating with a constant angular velocity $1/\tau_{r}$ \cite{ren2021synthesis}. 
Hence, the system is transverse contracting in region $\mathcal{C}_{\mathcal{T}}=\mathbb{R}^{2}\setminus\{\mathbf{0}\}$ \cite{manchester2014transverse}.

The phase variable of the Andronov-Hopf oscillator is used as the canonical system for rhythmic movement, i.e., $s_r(t)=\arctan( x_2(t)/x_1(t))$.
To generate a $T$-periodic function using $s_r$, $\tau_r$ is chosen to be $\tau_r=T/(2\pi)$.
Any positive values can be used for $\gamma$; a common choice is $\gamma=1$ to yield a unit circle limit cycle \cite{pham2009contraction}.

\begin{figure*}
  \includegraphics[trim={0.0cm 0.0cm 0.0cm 0.0cm}, width=0.99\textwidth, clip, page=1]{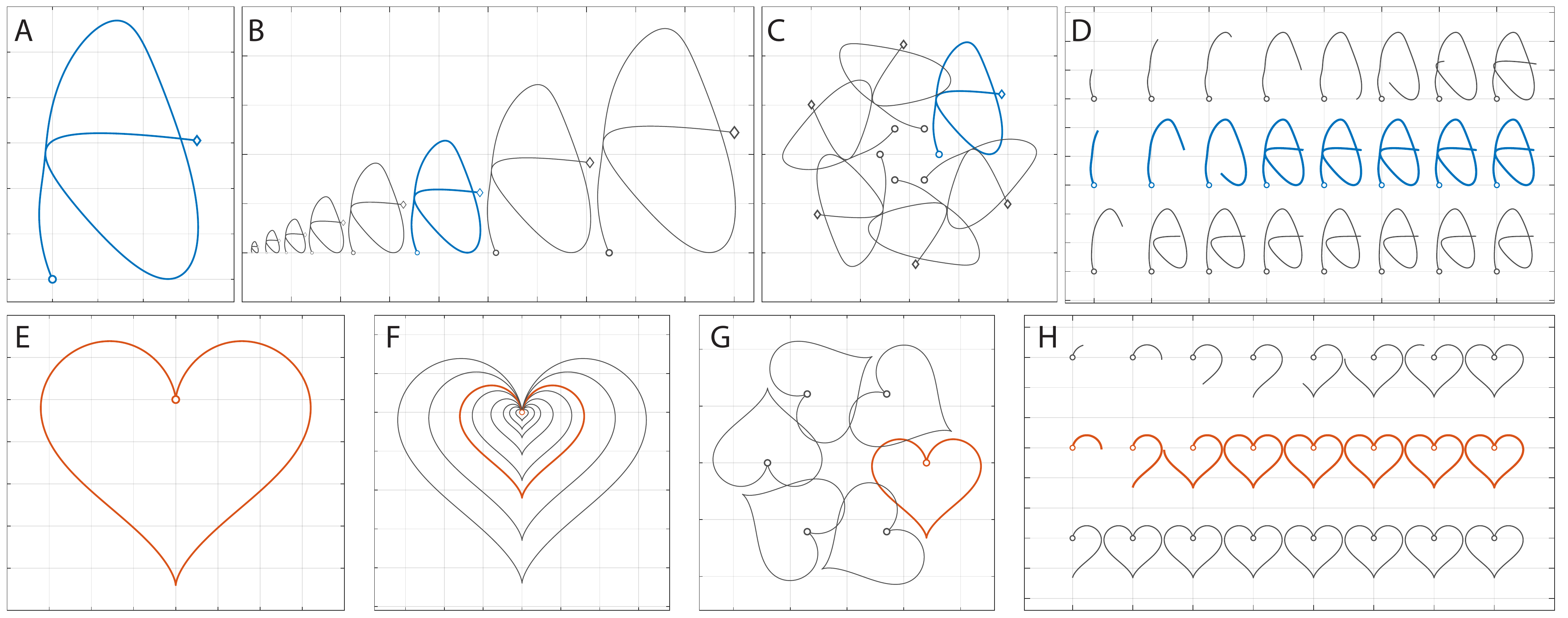}
  \caption{Dynamic movement primitives (DMP) for (A-D) discrete and (E-F) rhythmic movements and their (B, F) scaling, (C, G) rotational, and (D, H) temporal invariance properties. 
  Discrete and rhythmic movements are depicted with blue and orange colors, respectively. (A, E) Discrete and rhythmic DMPs for 
  letter A and heart, respectively. DMP parameters used for Imitation Learning: $\alpha_z=100$, $\beta_z=25$, $\alpha_s=1.0$, $N=50$, $P=1200$, $\tau_{d}^{(d)}=\tau_{r}^{(d)}=1.0$s, $T=2\pi$, $\mathbf{S}=\mathbb{I}_2$. (B, F) Spatial invariance property for discrete and rhythmic DMPs (Section \ref{subsubsec:spatial_temporal_invariance}). Values used for spatial scaling: $\kappa_s=0.1, 0.2, 0.3, 0.5, 0.8, 1.0, 1.5, 2.0$, ordered from left to right for discrete DMP (B), ordered from inside to outside for rhythmic DMP (F). (C, G) Rotational invariance property for discrete and rhythmic DMPs (Section \ref{subsubsec:rotational_invariance}). Spatial rotation matrices used for rotation: $\mathbf{R}=\text{rotz}(0^{\circ}),\text{rotz}(60^{\circ}), \text{rotz}(120^{\circ}), \text{rotz}(180^{\circ}), \text{rotz}(240^{\circ}), \text{rotz}(300^{\circ})$, ordered counterclockwise from the colored trajectories; $\text{rotz}$ denotes the rotation along $+Z$, which is positive out of the page. (D, H) Temporal invariance property for discrete and rhythmic DMPs (Section \ref{subsubsec:spatial_temporal_invariance}). Trajectories on top, middle, bottom rows depict trajectories generated by temporal scaling values of $\kappa_t=0.5, 1.0, 2.0$, respectively. Ordered from left to right columns, trajectories at time $t=[0.25, 0.50, 0.75, 1.0, 1.25, 1.5, 1.75, 2.0]\tau_{d}^{(d)}$ and $t=[0.25, 0.50, 0.75, 1.0, 1.25, 1.5, 1.75, 2.0]T$ are depicted for figure (D) and (H), respectively. For all figures, circle markers depict the start position for discrete and rhythmic DMPs; diamond markers depict the goal position of discrete DMP. For figures (B-D) and (E-G), blue and orange trajectories depict the trajectories in (A) and (B), respectively. }
  \label{fig1}
\end{figure*}

\subsubsection*{(Remark 3)}
The Andronov-Hopf oscillator is transverse contracting. 
Its dynamics 
(Eq.~\eqref{eq:canonical_system_andronov}) in polar coordinates, $r\equiv \sqrt{x_1^2 + x_2^2}$, $\theta\equiv \arctan(x_1/x_2)$ are given by:
\begin{equation}\label{eq:andronov_hopf_oscillator_polar}
    \begin{bmatrix}
        \dot{r} \\
        \dot{\theta}
    \end{bmatrix}
    =
    \begin{bmatrix}
        r ( \gamma^2 - r^2 ) \\
        1/\tau_{r}
    \end{bmatrix} 
    \equiv \mathbf{f}_{r\theta}(r)
\end{equation}
In this equation, $\mathbf{f}_{r\theta}$ is the vector field of the Andronov-Hopf oscillator in polar coordinates. 
Given an arbitrary positive constant $\epsilon\in\mathbb{R}_{>0}$, consider region $r\ge \epsilon$.
In this region, define a transverse contraction metric $\mathbf{M}_{\mathcal{T}}(r,\theta)\in\mathbb{R}^{2\times 2}$ which is symmetric and positive definite:
\begin{equation}
    \mathbf{M}_{\mathcal{T}}(r,\theta) = 
    \begin{bmatrix}
        1/r^2 & -\tau_r (\gamma^2-r^2)/r \\
        -\tau_r (\gamma^2-r^2)/r & m_{\theta\theta}(r,\theta)
    \end{bmatrix}
\end{equation}
The quantity $m_{\theta\theta}(r,\theta)\in\mathbb{R}_{>0}$ is chosen so that $m_{\theta\theta}(r,\theta) > (\gamma^2-r^2)\tau_r^2$ has a symmetric positive definite transverse contraction metric $\mathbf{M}_{\mathcal{T}}$.
Consequently, for $\delta \mathbf{x} = c[1,0]^{\top}$ with $c\in\mathbb{R}$, 
(1) $\delta \mathbf{x}^{\top}\mathbf{M}_{\mathcal{T}}\mathbf{f}_{r\theta}=0$, and (2) the transverse contraction condition (Eq.~\eqref{eq:transverse_contracting}) is satisfied with transverse contraction rate $\lambda_{\mathcal{T}} = 4\epsilon^2$. 
The transverse contraction region is $\mathcal{C}_{\mathcal{T}}=\{(r,\theta)\in\mathbb{R}^{2} ~|~ \epsilon \in \mathbb{R}_{>0}, r \ge \epsilon, \theta\in[0,2\pi)\}$. 

\subsection{Nonlinear Forcing Term}
A nonlinear forcing term, which takes the canonical system as a function argument, consists of a weighted sum of nonlinear basis functions.
As with the canonical system, different types of nonlinear forcing terms are used to generate discrete and rhythmic movements.

\subsubsection{For Discrete Movement}
A nonlinear forcing term for discrete movement, $\mathbf{F}_{d}:\mathbb{R}_{>0}\rightarrow \mathbb{R}^{n}$, is defined by a weighted sum of Gaussian functions:\footnote{While capital letters denote vectors, tuples, or arrays in this paper, an exception is made for the nonlinear forcing term to avoid confusion with the notation of nonlinear vector fields.}
\begin{align}\label{eq:DMP_nonlinear_force_discrete}
    \mathbf{F}_d(s_d(t)) &= \mathbf{W} \bm{\sigma}(s_d(t)) \nonumber \\ 
    \bm{\sigma}(s_d(t)) &= \frac{s_{d}(t)}{\sum_{i=1}^{N}\sigma_i(s_{d}(t))}
    \begin{bmatrix}
        \sigma_1(s_d(t)) \;\cdots\;  \sigma_N(s_d(t))
    \end{bmatrix}^{\top} \nonumber \\ 
    \sigma_{i}(s_{d}(t)) &= \exp\big\{ -h_i(s_{d}(t)-c_i)^2 \big\}     
\end{align}
$\mathbf{W}\in\mathbb{R}^{n\times N}$ is a weight matrix; $c_i, h_i\in\mathbb{R}$ are the center location and inverse width of the $i$-th Gaussian basis function $\sigma_{i}$, respectively; $N$ is the number of basis functions. 

To define $\mathbf{F}_d$, parameters $N$, $c_i$, $h_i$ are manually chosen \cite{ijspeert2013dynamical}; a common choice for these parameters is $c_i=\exp(-\alpha_s(i-1)/(N-1))$ for $i\in\{1,2,\cdots, N\}$, $h_{i}=1/(c_{i+1}-c_i)^2$ for $i\in\{1,2,\cdots, N-1\}$ and $h_{N-1}=h_{N}$ \cite{saveriano2023dynamic}.
On the other hand, the weight matrix $\mathbf{W}$ can be learned through various methods \cite{peters2008natural,kober2009learning,theodorou2010generalized, ijspeert2013dynamical}. Details for choosing $\mathbf{W}$ are deferred to Section \ref{subsec:imitation_learning}.

\subsubsection{For Rhythmic Movement}
A nonlinear forcing term for rhythmic movement, $\mathbf{F}_{r}:[0,2\pi)\rightarrow \mathbb{R}^{n}$, is defined by a weighted sum of von Mises functions:
\begin{align}\label{eq:DMP_nonlinear_force_rhythmic}
        \mathbf{F}_r(s_r(t)) &= \mathbf{W} \bm{\psi}(s_r(t)) \nonumber \\ 
        \bm{\psi}(s_r(t)) &= \frac{1}{\sum_{i=1}^{N}\psi_i(s_r(t))}
        \begin{bmatrix}
            \psi_1(s_r(t)) \cdots \psi_N(s_r(t))
        \end{bmatrix}^{\top} \nonumber \\
        \psi_{i}(s_{r}(t)) &= \exp\big\{  h_i (\cos(s_{r}(t)-c_i)-1) \big\}
\end{align}

As with the nonlinear forcing term for discrete movement, parameters $N$, $c_i$, $h_i$ for $\mathbf{F}_r$ are manually chosen \cite{ijspeert2013dynamical}; a common choice is $c_i=2\pi(i-1)/(N-1)$ and $h_i=2.5N$ for $i\in\{1,2,\cdots, N\}$ \cite{peternel2016adaptive,saveriano2023dynamic}.
Details for choosing $\mathbf{W}$ are deferred to Section \ref{subsec:imitation_learning}.

\begin{figure*}
    \centering
  \includegraphics[trim={0.0cm 0.0cm 0.0cm 0.0cm}, width=0.99\textwidth, clip, page=1]{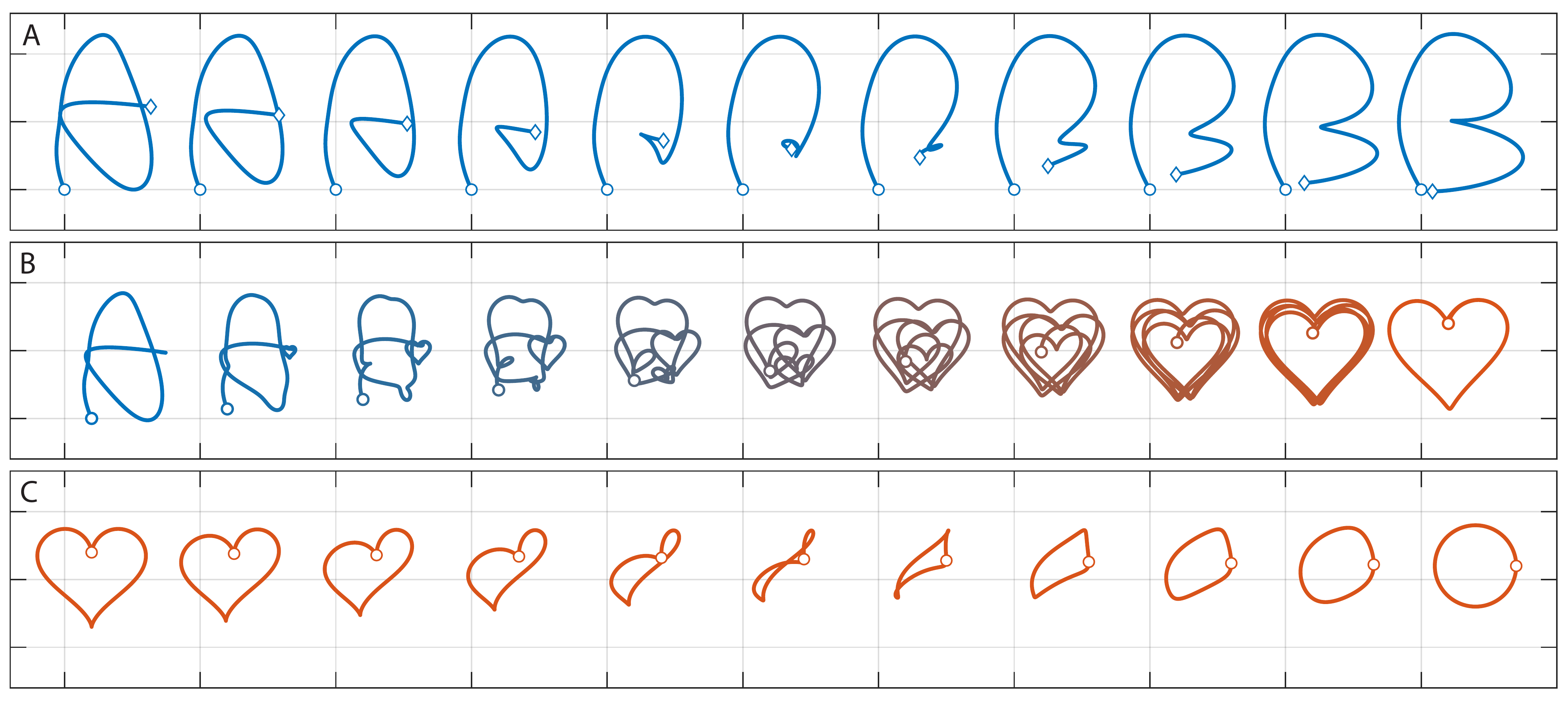}
  \caption{ Parallel combination of both discrete and rhythmic DMPs. (A) Parallel combination of discrete DMPs. For the combination, primitive inputs to generate the letters A and B were used. (B) Parallel combination of discrete and rhythmic DMPs. For the combination, primitive inputs to generate letter A (discrete) and heart (rhythmic) were used. (C) Parallel combination of rhythmic DMPs. For the combination, primitive inputs to generate heart and circle were used. For the 11 trajectories of each figure, the parameters used for parallel combination were $\alpha_1=[0.0,0.1,\cdots, 1.0]$ and $\alpha_2=1-\alpha_1$, ordered from left to right. DMP parameters used for Imitation Learning were identical to those in Figure \ref{fig1}. For all DMPs, the parameters for spatial scaling and rotation were $\kappa_{s}=1.0$, $\mathbf{R}=\mathbb{I}_2$. For temporal scaling, $\kappa_t=1.0$ was used for the trajectories in figure (A) and (C); for figure (B), $\kappa_t=0.25, 6.00$ were used for letter A (discrete) and heart (rhythmic), respectively. For all figures, circle markers depict the start position for discrete and rhythmic DMPs; diamond markers depict the goal position of discrete DMP.  }
  \label{fig2}
\end{figure*}

\subsection{Transformation System}
The nonlinear forcing term $\mathbf{F}_{d}$ (respectively $\mathbf{F}_{r})$ with canonical system $s_{d}$ (respectively $s_{r}$) as its function argument, is used as an input to the transformation system to generate discrete (respectively rhythmic) trajectories.
The transformation system for $\mathbb{R}^{n}$ consists of $n$ second-order linear systems with a scaled nonlinear forcing term $\mathbf{F}$:
\begin{equation}\label{eq:DMP_transformation_system}
    \begin{aligned}
        \tau \dot{\mathbf{y}}(t) &= \mathbf{z}(t) \\
        \tau \dot{\mathbf{z}}(t) &= \alpha_z \{ \beta_z ( \mathbf{g}-\mathbf{y}(t)) - \mathbf{z}(t)\} + \mathbf{S}\; \mathbf{F}(s(t))
    \end{aligned}    
\end{equation}
In these equations, $\mathbf{F}$ and its argument $s(t)$ denote ($\mathbf{F}_{d}$, $s_d$) or ($\mathbf{F}_{r}$, $s_r$) to generate discrete or rhythmic movements, respectively; 
$\mathbf{y}(t), \mathbf{z}(t)\in\mathbb{R}^{n}$ are the position and time-scaled velocity; $\alpha_z, \beta_z\in\mathbb{R}_{>0}$ are positive coefficients of the transformation system; $\tau\in\mathbb{R}_{>0}$ is the value used for the canonical system ($\tau_d$ for discrete movement, $\tau_r$ for rhythmic movement) (Section \ref{subsec:canonical_system}); $\mathbf{g}\in\mathbb{R}^{n}$ is a constant array to which $\mathbf{y}(t)$ converges when $\mathbf{F}\equiv \mathbf{0}$; $\mathbf{S}\in\mathbb{R}^{n\times n}$ is the scaling matrix for the nonlinear forcing term.

While any positive values of $\alpha_z$, $\beta_z$ can be used, usually the values of $\alpha_z$, $\beta_z$ are chosen such that if $\mathbf{F}=\mathbf{0}$, the transformation system is critically damped (i.e., has repeated eigenvalues) for $\tau=1$ (i.e., $\beta_z=\alpha_z/4$) \cite{ijspeert2013dynamical,nah2024robot}.

To generate trajectories $\mathbf{y}(t),\mathbf{z}(t)$, the initial conditions $\mathbf{y}(t=0)$, $\mathbf{z}(t=0)$ should be chosen and forward integrated. 
For this, appropriate choices of $\mathbf{S}$, $\mathbf{g}$, $\tau$ must be made. 
Details of choosing these parameters are discussed in Section \ref{subsec:generate_trajectory}.

\subsection{Imitation Learning}\label{subsec:imitation_learning}
Without the nonlinear forcing term, the trajectory for the transformation system $\mathbf{y}(t)$ follows the response of a stable second-order linear system which (exponentially) converges to $\mathbf{g}$.
To generate a wider range of movements for $\mathbf{y}(t)$ (e.g., rhythmic movements), the weight matrix $\mathbf{W}$ term can be learned through various methods \cite{kober2009learning, kober2013reinforcement}. 

One prominent approach is Imitation Learning \cite{schaal1999imitation,ijspeert2013dynamical,saveriano2023dynamic}, where the best-fit weights are learned from a demonstrated trajectory $\mathbf{y}^{(d)}(t)$. To conduct Imitation Learning, the position, velocity and acceleration of the demonstrated trajectory must be collected \cite{ijspeert2013dynamical,saveriano2023dynamic,nah2024robot}. 
Let the $P$ data points of these values be denoted by $\mathbf{y}^{(d)}(t_i), \dot{\mathbf{y}}^{(d)}(t_i), \ddot{\mathbf{y}}^{(d)}(t_i) \in\mathbb{R}^{n}$ for $i\in\{1, 2,\cdots, P\}$; for the position data, offset is added such that $\mathbf{y}^{(d)}(t_1)=\mathbf{0}$, i.e., the data points start at the origin.
With these $P$ data points, the weight matrix $\mathbf{W}$ is calculated by linear least-square regression \cite{saveriano2023dynamic}:
\begin{equation}
    \mathbf{W} = \mathbf{B}\mathbf{A}^{\top} ( \mathbf{A} \mathbf{A}^{\top})^{-1}
\end{equation}
For matrices $\mathbf{A}\in\mathbb{R}^{N\times P}$ and $\mathbf{B}\in\mathbb{R}^{n\times P}$, different values are used for discrete and rhythmic movements. 

\subsubsection{For Discrete Movement}
For discrete movements, matrices $\mathbf{A}$ and $\mathbf{B}$ are defined by:
\begin{align}\label{eq:Imitation_Learning_discrete}
    \mathbf{A} =&
    \begin{bmatrix}
        \mathbf{a}_{d}(t_{1}) & \mathbf{a}_{d}(t_{2}) & \cdots & \mathbf{a}_{d}(t_{P}) 
    \end{bmatrix} \nonumber \\ 
    \mathbf{B} =& 
    \begin{bmatrix}
        \mathbf{b}_{d}(t_{1}) & \mathbf{b}_{d}(t_{2}) & \cdots & \mathbf{b}_{d}(t_{P})
    \end{bmatrix} \nonumber \\  
    \mathbf{a}_{d}(t) =& s_{d}^{(d)}(t) \bm{\sigma}(s_d^{(d)}(t)) ~~~~~ s_d^{(d)}(t) = \exp\Big(-\frac{\alpha_s}{\tau_{d}^{(d)}}t\Big) \nonumber \\
    \mathbf{b}_{d}(t) =& (\tau_{d}^{(d)})^2 \ddot{\mathbf{y}}^{(d)}(t) + \alpha_z \tau_{d}^{(d)} \dot{\mathbf{y}}^{(d)}(t) \nonumber \\
    &+ \alpha_z \beta_z (\mathbf{y}^{(d)}(t)-\mathbf{g}_{d}^{(d)})
\end{align}
In these equations, $\tau_{d}^{(d)}\in\mathbb{R}_{>0}$ and $\mathbf{g}_{d}^{(d)}\in\mathbb{R}^{n}$ are the duration and goal location of the demonstrated trajectory, respectively, i.e., $\tau_{d}^{(d)}=t_P-t_1$ and $\mathbf{g}_{d}^{(d)}=\mathbf{y}^{(d)}(t_P)$.

\subsubsection{For Rhythmic Movement}
For rhythmic movements, matrices $\mathbf{A}$ and $\mathbf{B}$ are defined by:
\begin{align}\label{eq:Imitation_Learning_rhythmic}
    \mathbf{A} =&
    \begin{bmatrix}
        \mathbf{a}_{r}(t_{1}) & \mathbf{a}_{r}(t_{2}) & \cdots & \mathbf{a}_{r}(t_{P}) 
    \end{bmatrix} \nonumber \\
    \mathbf{B} =& 
    \begin{bmatrix}
        \mathbf{b}_{r}(t_{1}) & \mathbf{b}_{r}(t_{2}) & \cdots & \mathbf{b}_{r}(t_{P})
    \end{bmatrix} \nonumber \\  
    \mathbf{a}_{r}(t) =& \bm{\psi}(s_r^{(d)}(t)) \nonumber  ~~~~~ s_r^{(d)}(t) = \frac{t}{\tau_{r}^{(d)}} \mod{2\pi}  \\
    \mathbf{b}_{r}(t) =& (\tau_{r}^{(d)})^2 \ddot{\mathbf{y}}^{(d)}(t) + \alpha_z \tau_{r}^{(d)} \dot{\mathbf{y}}^{(d)}(t) \nonumber \\
    &+ \alpha_z \beta_z (\mathbf{y}^{(d)}(t)-\mathbf{g}_{r}^{(d)})
\end{align}
In these equations, $\tau_{r}^{(d)}\in\mathbb{R}_{>0}$ is $\tau_{r}^{(d)}=\frac{T^{(d)}}{2\pi}$ where $T^{(d)}=t_P-t_1$ is the period of the demonstrated trajectory. 
We assume the $P$ data points $\mathbf{y}^{(d)}(t_1)=\mathbf{0}, \mathbf{y}^{(d)}(t_2), \cdots, \mathbf{y}^{(d)}(t_P)$ comprise a closed-loop trajectory with an average position $\mathbf{g}_{r}^{(d)}=(\sum_{i=1}^{P}\mathbf{y}^{(d)}(t_i))/P$.

\begin{figure*}
    \centering
  \includegraphics[trim={0.0cm 0.0cm 0.0cm 0.0cm}, width=0.90\textwidth, clip, page=1]{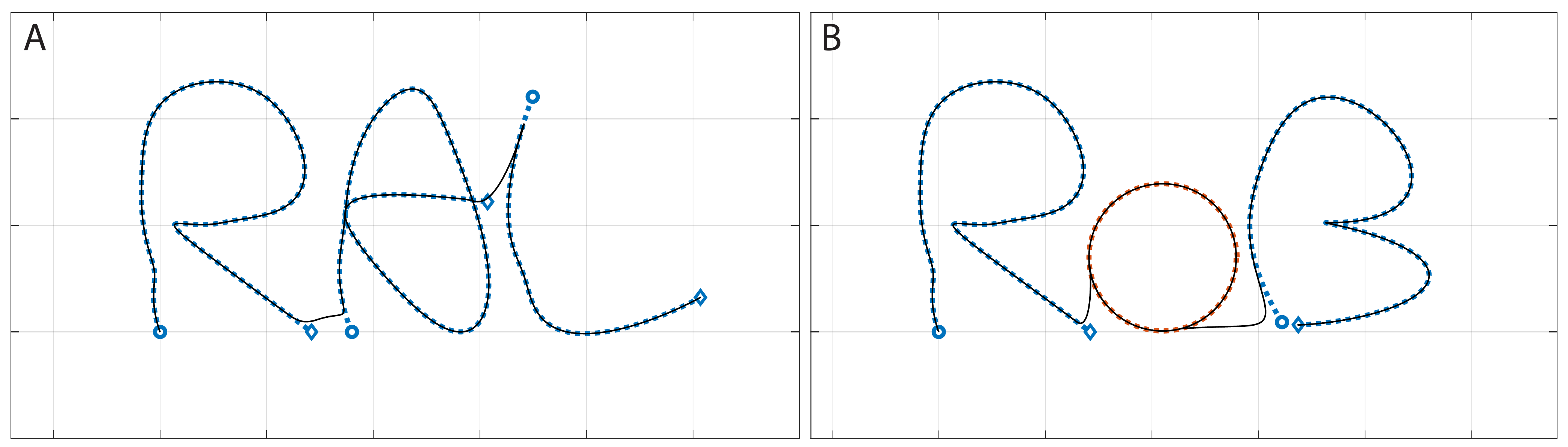}
  \caption{ Sequential combination of both discrete and rhythmic DMPs. (A) Sequential combination of three discrete DMPs, which generate Alphabets ``R'', ``A'' and ``L''. (B) Sequential combination of two discrete DMPs, which generate Alphabets ``R'' and ``B'', and one rhythmic DMP which generates a circular trajectory. DMP parameters used for Imitation Learning were identical to those in Figure \ref{fig1}. Black filled trajectories depict the result of the sequenced DMP; blue and orange dotted trajectories depict the trajectory of each individual discrete and rhythmic DMPs, respectively; circle and diamond markers depict the start and goal positions of discrete DMP, respectively. Time offsets for each movements (Eq.~\eqref{eq:sequential_combination_primitives}): (A) $t_{off,1}=0.0$s, $t_{off,2}=0.969$s, $t_{off,3}=1.939$s; (B) $t_{off,1}=0.0$s, $t_{off,2}=0.970$s, $t_{off,3}=4.499$s. The corresponding parameters for the activation function (Eq.~\eqref{eq:activation_function}) were selected to achieve a smooth transition to the subsequent trajectory.}
  \label{fig3}
\end{figure*}

\subsection{Generating the Trajectory}\label{subsec:generate_trajectory}
Once the weight matrix $\mathbf{W}$ of the nonlinear forcing term is learned via Imitation Learning, the transformation system is  integrated
using the values of $\mathbf{S}$, $\mathbf{g}$, and $\tau$ to determine the shape of $\mathbf{y}(t)$. 
For both discrete and rhythmic DMPs, if one uses $\mathbf{S}=\mathbb{I}_n$, $\mathbf{g}=\mathbf{g}^{(d)}$ (i.e., $\mathbf{g}_d^{(d)}$ for discrete, $\mathbf{g}_r^{(d)}$ for rhythmic), $\tau=\tau^{(d)}$ (i.e., $\tau_d^{(d)}$ for discrete, $\tau_r^{(d)}$ for rhythmic), and initial conditions  $\mathbf{y}(t=0)=\mathbf{0}$, $\mathbf{z}(t=0)=\mathbf{0}$ then $\mathbf{y}(t)\rightarrow \mathbf{y}^{(d)}(t)$, i.e., the transformation system generates a trajectory provided by demonstration (Figure \ref{fig1}A, \ref{fig1}E).

\subsubsection{Spatial and Temporal Invariance Properties}\label{subsubsec:spatial_temporal_invariance}
To spatially and temporally scale the learned trajectory $\mathbf{y}^{(d)}(t)$, the following parameters are chosen: $\mathbf{S}=\kappa_s\mathbb{I}_n$, $\mathbf{y}(0)=\mathbf{y}_0$, $\mathbf{z}(0)=\mathbf{0}$, $\mathbf{g}=\mathbf{y}_0+\kappa_s\mathbf{g}^{(d)}$, $\tau=\kappa_t \tau^{(d)}$ where $\kappa_s,\kappa_t \in\mathbb{R}_{>0}$ are the spatial and temporal scaling factors, respectively.
With these parameters, $\mathbf{y}(t)\rightarrow \mathbf{y}_0+\kappa_s \mathbf{y}^{(d)}(\kappa_t t)$ (Figure \ref{fig1}B, \ref{fig1}D, \ref{fig1}F, \ref{fig1}H).

\subsubsection{Rotational Invariance Property}\label{subsubsec:rotational_invariance}
For $n=3$, one can also spatially rotate the demonstrated trajectory. Given an arbitrary spatial rotation matrix $\mathbf{R}\in\text{SO}(3)$, the following parameters are chosen: $\mathbf{S}=\kappa_s\mathbf{R}$, $\mathbf{y}(0)=\mathbf{y}_0$, $\mathbf{z}(0)=\mathbf{0}$, $\mathbf{g}=\mathbf{y}_0+\kappa_s\mathbf{R}\mathbf{g}^{(d)}$, $\tau=\kappa_t \tau^{(d)}$.
With these parameters, $\mathbf{y}(t)\rightarrow \mathbf{y}_0+\kappa_s \mathbf{R}\mathbf{y}^{(d)}(\kappa_t t)$ (Figure \ref{fig1}C, \ref{fig1}G).

\section{Contribution: Combining Movement Primitives with Contraction Theory}
In this Section, we show how both discrete and rhythmic DMPs can be combined to generate parallel and sequential combinations of movement primitives.

\subsection{Definition of \revisionone{Movement} Primitive Input}
Consider a nonlinear forcing term $\mathbf{F}(s(t))$ (either for discrete or rhythmic DMP), where the weights $\mathbf{W}$ are learned via Imitation Learning (Section \ref{subsec:imitation_learning}) with parameters $\alpha_z, \beta_z, \tau^{(d)}$, $\mathbf{g}^{(d)}$ (and additionally $\alpha_s$ for discrete DMP). 

A \revisionone{movement} primitive input, $\mathbf{p}(\mathbf{x}(t))\in\mathbb{R}^{n}$ is defined by:
\begin{equation}\label{eq:definition_kin_prim}
    \mathbf{p}(\mathbf{x}(t)) = \mathbf{F}(s^{(d)}(t)) + \alpha_z \beta_z \mathbf{g}^{(d)}
\end{equation}
where:
\begin{equation}
    \mathbf{F}(s^{(d)}(t)) = 
        \begin{cases}
          \mathbf{W} \bm{\sigma}(s_d^{(d)}(t))  & \text{for Discrete DMP} \\
          \mathbf{W} \bm{\psi}(s_r^{(d)}(t)) & \text{for Rhythmic DMP}
        \end{cases}
\end{equation}
With this definition 
for discrete and rhythmic movements, $\mathbf{p}(\mathbf{x})$ is used as an input to a transformation system with $\mathbf{g}=\mathbf{0}$ (Eq.~\eqref{eq:DMP_transformation_system}):
\begin{equation}\label{eq:definition_of_movement_primitives}
    (\tau^{(d)})^2 \ddot{\mathbf{y}}(t) + \tau^{(d)}\alpha_z \dot{\mathbf{y}}(t) + \alpha_z\beta_z \mathbf{y}(t) = \mathbf{p}(\mathbf{x})
\end{equation}
Given initial conditions $\mathbf{y}(t=0)=\mathbf{0}$, $\mathbf{z}(t=0)=\mathbf{0}$, $\mathbf{y}(t)\rightarrow \mathbf{y}^{(d)}(t)$ (Section \ref{subsec:imitation_learning}) (Figure \ref{fig1}A, \ref{fig1}E).
Primitive $\mathbf{p}(\mathbf{x})$ generates $\mathbf{y}^{(d)}(t)$ as provided by demonstration.

\subsubsection*{(Remark 4)}
The stability of this approach is proven 
using Contraction Theory \cite{perk2006motion}, \cite{wensing2017sparse}.
A transformation system itself is a (globally) contracting system.
\revisionone{Movement} primitives $\mathbf{p}(\mathbf{x})$ for discrete and rhythmic movements are contracting and transverse contracting systems, respectively. 
For discrete (respectively rhythmic) DMP, the transformation system with input $\mathbf{p}(\mathbf{x})$ is a hierarchical combination of contracting systems (respectively contracting and transverse contracting systems), which thereby results in a contracting (respectively transverse contracting) system.

\subsubsection*{(Remark 5)}
As discussed in Section \ref{subsec:generate_trajectory}, spatial and temporal scaling and rotation 
can be achieved by scaling the primitive inputs. 
In detail, using $\kappa_{s}\mathbf{R} \mathbf{p}(\mathbf{x}(\kappa_t t))\equiv \kappa_s \mathbf{R}\{\mathbf{F}(s^{(d)}(\kappa_t t)) + \alpha_z \beta_z \mathbf{g}^{(d)}\}$, as an input to Eq.~\eqref{eq:definition_of_movement_primitives}, given initial conditions $\mathbf{y}(t=0)=\mathbf{y}_0$, $\mathbf{z}(t=0)=\mathbf{0}$, $\mathbf{y}(t)\rightarrow \mathbf{y}_0 + \kappa_{s}\mathbf{R}\mathbf{y}^{(d)}(\kappa_t t)$ (Figure \ref{fig1}B-D, \ref{fig1}F-H). 
Moreover, an offset $\mathbf{y}_{off}\in\mathbb{R}^{n}$ can be added to the trajectories by adding $\alpha_z\beta_z\mathbf{y}_{off}$ to the \revisionone{movement} primitive input. 
Then, for initial conditions $\mathbf{y}(t=0)=\mathbf{y}_0+\mathbf{y}_{off}$, $\mathbf{z}(t=0)=\mathbf{0}$, $\mathbf{y}(t)\rightarrow (\mathbf{y}_0 + \mathbf{y}_{off}) + \kappa_{s}\mathbf{R}\mathbf{y}^{(d)}(\kappa_t t)$.

\subsection{Combining \revisionone{Movement} Primitives}
Parallel and sequential combinations of both discrete and rhythmic DMPs can be achieved. 

\subsubsection{Application 1: Parallel Combination}
Consider two primitive inputs, $\mathbf{p}_1(\mathbf{x}_1(t))$ and $\mathbf{p}_2(\mathbf{x}_2(t))$, which can represent either discrete or rhythmic DMPs and may be spatially, temporally scaled, or rotated.
A parallel combination of these two movement primitives can be achieved by a parallel combination of the two primitive inputs (Section \ref{subsubsec:parallel_combination}):
\begin{equation}
    \begin{gathered}
         \alpha_1, \alpha_2 \ge 0: \mathbf{p}(\mathbf{x}) = \alpha_1\mathbf{p}_1(\mathbf{x}_1(t)) +  \alpha_2\mathbf{p}_2(\mathbf{x}_2(t)) \\
    \end{gathered}    
\end{equation}
And recursively, any number of primitives can be parallel combined. 

Figure \ref{fig2} illustrates examples of parallel combination of discrete DMPs (Figure \ref{fig2}A), discrete and rhythmic DMPs (Figure \ref{fig2}B), and rhythmic DMPs (Figure \ref{fig2}C).

\subsubsection*{(Remark 6)} 
\revisionone{One promising robotic application} of the presented method is motion planning for tasks which involve combinations of both discrete and rhythmic movements, e.g., polishing, grinding, peg-in-hole assembly tasks \cite{lachner2024divide,nah2024modularEDA}. 
\revisionone{As each movement can be spatially and temporally scaled or rotated, flexible and versatile motion planning can be achieved. For instance, in a polishing task, temporal scaling can adjust the speed of a rhythmic polishing motion, slowing it down for thorough polishing or speeding it up for rough surface treatment. Similarly, spatial scaling and rotation can modify the range of movement to adapt to varying surface sizes and orientations. 
Note that spatial or temporal scaling can be achieved \textit{by changing a single scalar variable}, thereby simplifying robot motion programming. }

\subsubsection{Application 2: Sequential Combination}
Consider two primitive inputs, $\mathbf{p}_1(\mathbf{x}_1(t))$ and $\mathbf{p}_2(\mathbf{x}_2(t))$, which can represent either discrete or rhythmic DMPs and may be spatially, temporally scaled, or rotated with a position offset. 
A sequential combination of these two movement primitives can be achieved by using time-varying weights, $\alpha_1(t)$, $\alpha_2(t)$ with time offsets of each primitive input (Section \ref{subsubsec:sequential_combination}):
\begin{equation}\label{eq:sequential_combination_primitives}
    \begin{gathered}
         \alpha_1(t), \alpha_2(t) \ge 0: \mathbf{p}(\mathbf{x}) = \\
          \alpha_1(t)\mathbf{p}_1(\mathbf{x}_1(t-t_{1,off})) +  \alpha_2(t)\mathbf{p}_2(\mathbf{x}_2(t-t_{2,off})) 
    \end{gathered}    
\end{equation}
In this equation, $t_{1,off}, t_{2,off} \in \mathbb{R}_{>0}$ are the time offsets. 
Recursively, any number of primitives can be sequentially combined.

Figure \ref{fig3} illustrates examples of sequential combination of both discrete and rhythmic DMPs. 
The time-varying weighting function used for this example was:
\begin{equation}\label{eq:activation_function}
    \alpha_i(t) = 
    \begin{cases}
        0 & \text{$0 \le t < t_1$}\\
        3(\frac{t-t_1}{t_2-t_1})^2 - 2(\frac{t-t_1}{t_2-t_1})^3  & \text{$t_1 \le t < t_2$}  \\
        1 & \text{$t_2 \le t < t_3 $} \\
        1 - 3(\frac{t-t_3}{t_4-t_3})^2 + 2(\frac{t-t_3}{t_4-t_3})^3  & \text{$t_3 \le t < t_4$}  \\
        0 & \text{$t_4 \le t$}
    \end{cases}
\end{equation}
where $i\in \{1,2,\cdots, N_p\}$ and $t_1, t_2, t_3, t_4\in\mathbb{R}_{>0}$. 

Compared to prior approaches \cite{khansari2011learning,figueroa2022locally,billard2022learning,burridge1999sequential,majumdar2017funnel} which account for sequential combinations of discrete movements, the presented approach achieves a sequential combination of both discrete and rhythmic movements \textit{while preserving} the spatial, temporal, and rotational invariance properties of DMP (Section \ref{subsec:generate_trajectory}). 

\subsubsection*{(Remark 7)} 
This approach can be used for any robotic application involving a sequence of trajectories, e.g., planning of robot end-effector motion \cite{burridge1999sequential,saveriano2019merging}, or unmanned aerial vehicle trajectories (UAV) \cite{majumdar2017funnel}. \revisionone{The approach expands prior methods by incorporating rhythmic movements.}

\subsubsection{Application 3: Parallel and Sequential Combinations}
\begin{figure}[H]
    \centering
  \includegraphics[trim={0.0cm 0.0cm 0.0cm 0.0cm}, width=0.96\columnwidth, clip, page=1]{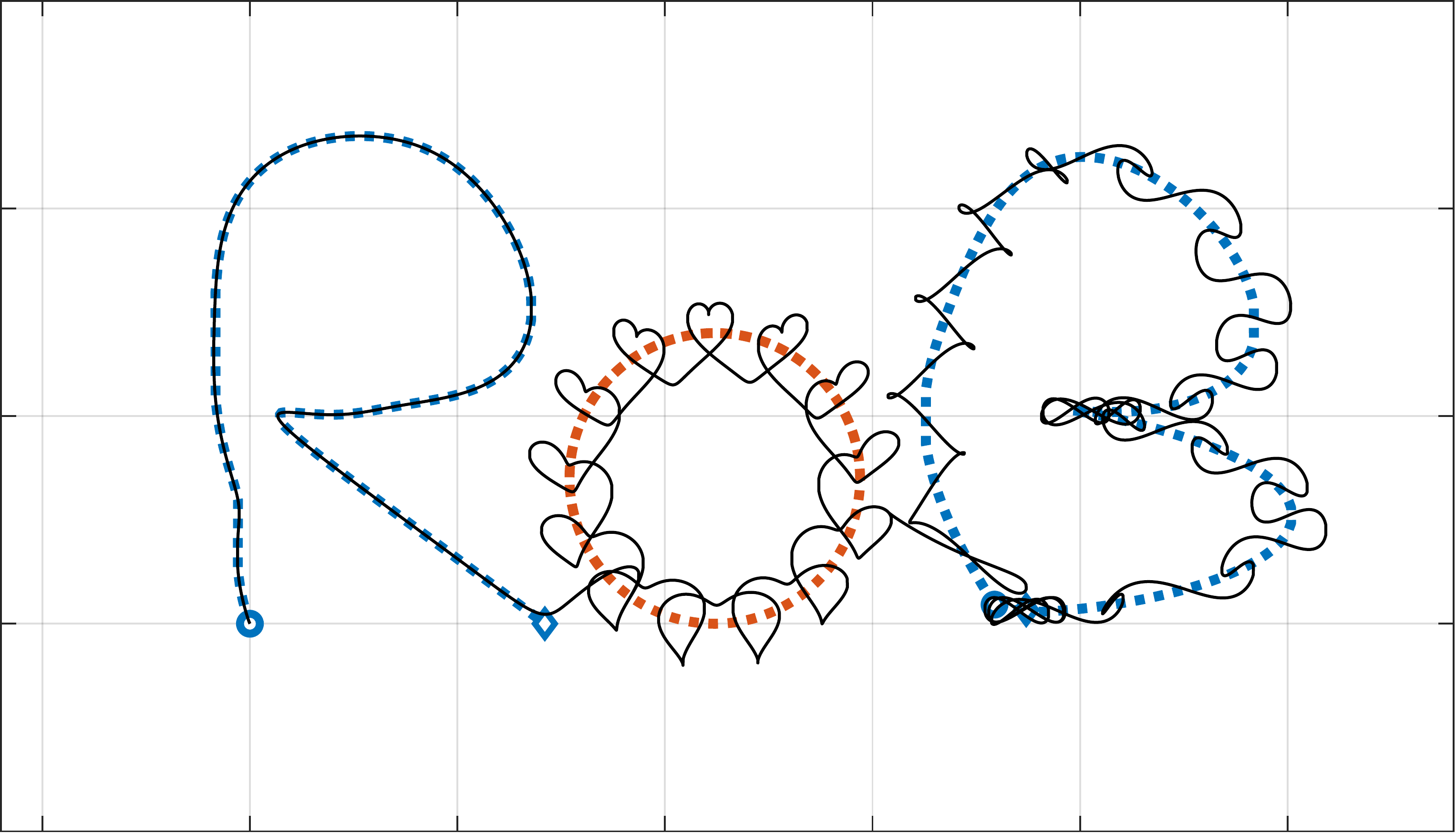}
  \caption{Parallel and sequential combinations of both discrete and rhythmic DMPs. On the sequenced trajectory used in Figure \ref{fig3}B, additional rhythmic DMPs were parallel combined. For the circular trajectory and letter B, rhythmic DMPs for heart (Figure \ref{fig2}B) and infinity symbol were parallel combined, respectively. }
  \label{fig4}
\end{figure}

Parallel and sequential combinations of discrete and rhythmic DMPs can be achieved. Figure \ref{fig4} illustrates an example.

\subsubsection*{(Remark 8)} 
\revisionone{The presented approach enables both parallel and sequential combinations of discrete and rhythmic movements, with independent modulation of each movement. 
This provides a divide-and-conquer strategy for robot programming---programmers can break down complex robot applications into simpler, manageable sub-tasks, each  associated with a (combination of) movement primitive(s), which can later be combined.
The ability to independently modulate each movement further simplifies programming for motion planning, allowing robot programmers to refine individual movements without the need to relearn the whole movement from scratch. 
This modular framework not only improves flexibility but also accelerates development by allowing the reuse and refinement of existing components, making the approach well-suited for diverse and complex motion planning challenges in robotics. }

\section{Discussion, Future Work and Conclusion}
This paper introduced a modular framework to combine discrete and rhythmic DMPs. 
The method preserves the spatial, temporal, and rotational invariance properties of DMP. 
Methods for parallel and/or sequential combinations of both discrete and rhythmic DMPs were presented and demonstrated through simulation examples.
Stability \revisionone{analysis} 
was provided using Contraction Theory, which enabled incorporation of rhythmic movements 
with stability guarantees. 
\revisionone{As a result}, robust and adaptable motion planning \revisionone{is achieved, allowing complex movements to be constructed by combining simpler movement primitives. }

Note that the combination of \revisionone{movement} primitive inputs is achieved \textit{under the same contraction metric} (Section \ref{subsubsec:parallel_combination}). The weights of each movement primitive must be learned under identical DMP parameters. 
While this constraint may appear to limit the method's flexibility, it facilitates additional applications, e.g., movement identification \cite{ijspeert2013dynamical}. 

As the approach achieves modular motion planning based on linear superposition of \revisionone{movement} primitive inputs, the system can be reactive (or adaptive) in changing environments. For instance, obstacle avoidance methods using an additional coupling term \cite{hoffmann2009biologically} can be seamlessly incorporated. 
Temporal coupling terms can be added to the canonical system to adapt to moving goals for discrete DMP \cite{ijspeert2013dynamical,koutras2020dynamic}, or to switch between different rhythmic patterns for rhythmic DMP (e.g., changing gait patterns in different terrain). 

Future work could explore actual robot implementation 
or an extension to different manifolds, e.g., the $\text{SO}(3)$ manifold for spatial orientation.



\bibliographystyle{IEEEtran}
\bibliography{references.bib}

\begin{thebibliography}{10}
\providecommand{\url}[1]{#1}
\csname url@samestyle\endcsname
\providecommand{\newblock}{\relax}
\providecommand{\bibinfo}[2]{#2}
\providecommand{\BIBentrySTDinterwordspacing}{\spaceskip=0pt\relax}
\providecommand{\BIBentryALTinterwordstretchfactor}{4}
\providecommand{\BIBentryALTinterwordspacing}{\spaceskip=\fontdimen2\font plus
\BIBentryALTinterwordstretchfactor\fontdimen3\font minus \fontdimen4\font\relax}
\providecommand{\BIBforeignlanguage}[2]{{%
\expandafter\ifx\csname l@#1\endcsname\relax
\typeout{** WARNING: IEEEtran.bst: No hyphenation pattern has been}%
\typeout{** loaded for the language `#1'. Using the pattern for}%
\typeout{** the default language instead.}%
\else
\language=\csname l@#1\endcsname
\fi
#2}}
\providecommand{\BIBdecl}{\relax}
\BIBdecl

\bibitem{williamson1999robot}
M.~M. Williamson, ``Robot arm control exploiting natural dynamics,'' Ph.D. dissertation, Massachusetts Institute of Technology, 1999.

\bibitem{ijspeert2002learning}
A.~Ijspeert, J.~Nakanishi, and S.~Schaal, ``Learning attractor landscapes for learning motor primitives,'' \emph{Advances in neural information processing systems}, vol.~15, 2002.

\bibitem{schaal2007dynamics}
S.~Schaal, P.~Mohajerian, and A.~Ijspeert, ``Dynamics systems vs. optimal control—a unifying view,'' \emph{Progress in brain research}, vol. 165, pp. 425--445, 2007.

\bibitem{righetti2008pattern}
L.~Righetti and A.~J. Ijspeert, ``Pattern generators with sensory feedback for the control of quadruped locomotion,'' in \emph{2008 IEEE International Conference on Robotics and Automation}.\hskip 1em plus 0.5em minus 0.4em\relax IEEE, 2008, pp. 819--824.

\bibitem{pastor2009learning}
P.~Pastor, H.~Hoffmann, T.~Asfour, and S.~Schaal, ``Learning and generalization of motor skills by learning from demonstration,'' in \emph{2009 IEEE International Conference on Robotics and Automation}.\hskip 1em plus 0.5em minus 0.4em\relax IEEE, 2009, pp. 763--768.

\bibitem{hogan2012dynamic}
N.~Hogan and D.~Sternad, ``Dynamic primitives of motor behavior,'' \emph{Biological Cybernetics}, vol. 106, no. 11-12, pp. 727--739, 2012.

\bibitem{ijspeert2013dynamical}
A.~J. Ijspeert, J.~Nakanishi, H.~Hoffmann, P.~Pastor, and S.~Schaal, ``Dynamical movement primitives: learning attractor models for motor behaviors,'' \emph{Neural Computation}, vol.~25, no.~2, pp. 328--373, 2013.

\bibitem{koutras2020correct}
L.~Koutras and Z.~Doulgeri, ``A correct formulation for the orientation dynamic movement primitives for robot control in the cartesian space,'' in \emph{Conference on robot learning}.\hskip 1em plus 0.5em minus 0.4em\relax PMLR, 2020, pp. 293--302.

\bibitem{khadivar2021learning}
F.~Khadivar, I.~Lauzana, and A.~Billard, ``Learning dynamical systems with bifurcations,'' \emph{Robotics and Autonomous Systems}, vol. 136, p. 103700, 2021.

\bibitem{figueroa2022locally}
N.~Figueroa and A.~Billard, ``Locally active globally stable dynamical systems: Theory, learning, and experiments,'' \emph{The International Journal of Robotics Research}, vol.~41, no.~3, pp. 312--347, 2022.

\bibitem{billard2022learning}
A.~Billard, S.~Mirrazavi, and N.~Figueroa, \emph{Learning for adaptive and reactive robot control: a dynamical systems approach}.\hskip 1em plus 0.5em minus 0.4em\relax Mit Press, 2022.

\bibitem{saveriano2023dynamic}
M.~Saveriano, F.~J. Abu-Dakka, A.~Kramberger, and L.~Peternel, ``Dynamic movement primitives in robotics: A tutorial survey,'' \emph{The International Journal of Robotics Research}, vol.~42, no.~13, pp. 1133--1184, 2023.

\bibitem{nah2024robot}
M.~C. Nah, J.~Lachner, and N.~Hogan, ``Robot control based on motor primitives: A comparison of two approaches,'' \emph{The International Journal of Robotics Research}, p. 02783649241258782, 2024.

\bibitem{abu2024unified}
F.~J. Abu-Dakka, M.~Saveriano, and V.~Kyrki, ``A unified formulation of geometry-aware discrete dynamic movement primitives,'' \emph{Neurocomputing}, p. 128056, 2024.

\bibitem{khansari2011learning}
S.~M. Khansari-Zadeh and A.~Billard, ``Learning stable nonlinear dynamical systems with gaussian mixture models,'' \emph{IEEE Transactions on Robotics}, vol.~27, no.~5, pp. 943--957, 2011.

\bibitem{koutras2020novel}
L.~Koutras and Z.~Doulgeri, ``A novel dmp formulation for global and frame independent spatial scaling in the task space,'' in \emph{2020 29th IEEE International Conference on Robot and Human Interactive Communication (RO-MAN)}.\hskip 1em plus 0.5em minus 0.4em\relax IEEE, 2020, pp. 727--732.

\bibitem{abu2024learning}
F.~Abu-Dakka, M.~Saveriano, and L.~Peternel, ``Learning periodic skills for robotic manipulation: Insights on orientation and impedance,'' \emph{Robotics and Autonomous Systems}, p. 104763, 2024.

\bibitem{hoffmann2009biologically}
H.~Hoffmann, P.~Pastor, D.-H. Park, and S.~Schaal, ``Biologically-inspired dynamical systems for movement generation: Automatic real-time goal adaptation and obstacle avoidance,'' in \emph{2009 IEEE international conference on robotics and automation}.\hskip 1em plus 0.5em minus 0.4em\relax IEEE, 2009, pp. 2587--2592.

\bibitem{khansari2012dynamical}
S.~M. Khansari-Zadeh and A.~Billard, ``A dynamical system approach to realtime obstacle avoidance,'' \emph{Autonomous Robots}, vol.~32, pp. 433--454, 2012.

\bibitem{peters2008natural}
J.~Peters and S.~Schaal, ``Natural actor-critic,'' \emph{Neurocomputing}, vol.~71, no. 7-9, pp. 1180--1190, 2008.

\bibitem{kober2009learning}
J.~Kober and J.~Peters, ``Learning motor primitives for robotics,'' in \emph{2009 IEEE International Conference on Robotics and Automation}.\hskip 1em plus 0.5em minus 0.4em\relax IEEE, 2009, pp. 2112--2118.

\bibitem{zhou2017task}
Y.~Zhou and T.~Asfour, ``Task-oriented generalization of dynamic movement primitive,'' in \emph{2017 IEEE/RSJ International Conference on Intelligent Robots and Systems (IROS)}.\hskip 1em plus 0.5em minus 0.4em\relax IEEE, 2017, pp. 3202--3209.

\bibitem{billard2019trends}
A.~Billard and D.~Kragic, ``Trends and challenges in robot manipulation,'' \emph{Science}, vol. 364, no. 6446, p. eaat8414, 2019.

\bibitem{huber2022fast}
L.~Huber, J.-J. Slotine, and A.~Billard, ``Fast obstacle avoidance based on real-time sensing,'' \emph{IEEE Robotics and Automation Letters}, vol.~8, no.~3, pp. 1375--1382, 2022.

\bibitem{lachner2024divide}
J.~Lachner, F.~Tessari, A.~M. West~Jr, M.~C. Nah, and N.~Hogan, ``Divide et impera: Learning impedance families for peg-in-hole assembly,'' \emph{arXiv preprint arXiv:2410.01054}, 2024.

\bibitem{strogatz2018nonlinear}
S.~H. Strogatz, \emph{Nonlinear dynamics and chaos: with applications to physics, biology, chemistry, and engineering}.\hskip 1em plus 0.5em minus 0.4em\relax CRC press, 2018.

\bibitem{burridge1999sequential}
R.~R. Burridge, A.~A. Rizzi, and D.~E. Koditschek, ``Sequential composition of dynamically dexterous robot behaviors,'' \emph{The International Journal of Robotics Research}, vol.~18, no.~6, pp. 534--555, 1999.

\bibitem{majumdar2017funnel}
A.~Majumdar and R.~Tedrake, ``Funnel libraries for real-time robust feedback motion planning,'' \emph{The International Journal of Robotics Research}, vol.~36, no.~8, pp. 947--982, 2017.

\bibitem{degallier2006movement}
S.~Degallier, C.~P. Santos, L.~Righetti, and A.~Ijspeert, ``Movement generation using dynamical systems: a humanoid robot performing a drumming task,'' in \emph{2006 6th IEEE-RAS International Conference on Humanoid Robots}.\hskip 1em plus 0.5em minus 0.4em\relax IEEE, 2006, pp. 512--517.

\bibitem{ernesti2012encoding}
J.~Ernesti, L.~Righetti, M.~Do, T.~Asfour, and S.~Schaal, ``Encoding of periodic and their transient motions by a single dynamic movement primitive,'' in \emph{2012 12th IEEE-RAS international conference on humanoid robots (humanoids 2012)}.\hskip 1em plus 0.5em minus 0.4em\relax IEEE, 2012, pp. 57--64.

\bibitem{lohmiller1998contraction}
W.~Lohmiller and J.-J.~E. Slotine, ``On contraction analysis for non-linear systems,'' \emph{Automatica}, vol.~34, no.~6, pp. 683--696, 1998.

\bibitem{slotine2003modular}
J.-J.~E. Slotine, ``Modular stability tools for distributed computation and control,'' \emph{International Journal of Adaptive Control and Signal Processing}, vol.~17, no.~6, pp. 397--416, 2003.

\bibitem{perk2006motion}
B.~E. Perk and J.-J.~E. Slotine, ``Motion primitives for robotic flight control,'' \emph{arXiv preprint cs/0609140}, 2006.

\bibitem{wensing2017sparse}
P.~M. Wensing and J.-J. Slotine, ``Sparse control for dynamic movement primitives,'' \emph{IFAC-PapersOnLine}, vol.~50, no.~1, pp. 10\,114--10\,121, 2017.

\bibitem{slotine2001modularity}
J.-J. Slotine and W.~Lohmiller, ``Modularity, evolution, and the binding problem: a view from stability theory,'' \emph{Neural networks}, vol.~14, no.~2, pp. 137--145, 2001.

\bibitem{tsukamoto2021contraction}
H.~Tsukamoto, S.-J. Chung, and J.-J.~E. Slotine, ``Contraction theory for nonlinear stability analysis and learning-based control: A tutorial overview,'' \emph{Annual Reviews in Control}, vol.~52, pp. 135--169, 2021.

\bibitem{bullo2022contraction}
F.~Bullo, \emph{Contraction theory for dynamical systems}.\hskip 1em plus 0.5em minus 0.4em\relax Francesco Bullo, 2022.

\bibitem{simpson2014contraction}
J.~W. Simpson-Porco and F.~Bullo, ``Contraction theory on riemannian manifolds,'' \emph{Systems \& Control Letters}, vol.~65, pp. 74--80, 2014.

\bibitem{manchester2014transverse}
I.~R. Manchester and J.-J.~E. Slotine, ``Transverse contraction criteria for existence, stability, and robustness of a limit cycle,'' \emph{Systems \& Control Letters}, vol.~63, pp. 32--38, 2014.

\bibitem{wang2005partial}
W.~Wang and J.-J.~E. Slotine, ``On partial contraction analysis for coupled nonlinear oscillators,'' \emph{Biological cybernetics}, vol.~92, no.~1, pp. 38--53, 2005.

\bibitem{ijspeert2002movement}
A.~J. Ijspeert, J.~Nakanishi, and S.~Schaal, ``Movement imitation with nonlinear dynamical systems in humanoid robots,'' in \emph{Proceedings 2002 IEEE International Conference on Robotics and Automation (Cat. No. 02CH37292)}, vol.~2.\hskip 1em plus 0.5em minus 0.4em\relax IEEE, 2002, pp. 1398--1403.

\bibitem{schaal2003control}
S.~Schaal, J.~Peters, J.~Nakanishi, and A.~Ijspeert, ``Control, planning, learning, and imitation with dynamic movement primitives,'' in \emph{Workshop on Bilateral Paradigms on Humans and Humanoids, IEEE International Conference on Intelligent Robots and Systems}, 2003, pp. 1--21.

\bibitem{ren2021synthesis}
K.~Ren, \emph{Synthesis of Linear Distributed Control and Coupled Oscillators with Multiple Limit Cycles}.\hskip 1em plus 0.5em minus 0.4em\relax University of California, Los Angeles, 2021.

\bibitem{pham2009contraction}
Q.-C. Pham, N.~Tabareau, and J.-J. Slotine, ``A contraction theory approach to stochastic incremental stability,'' \emph{IEEE Transactions on Automatic Control}, vol.~54, no.~4, pp. 816--820, 2009.

\bibitem{theodorou2010generalized}
E.~Theodorou, J.~Buchli, and S.~Schaal, ``A generalized path integral control approach to reinforcement learning,'' \emph{The Journal of Machine Learning Research}, vol.~11, pp. 3137--3181, 2010.

\bibitem{peternel2016adaptive}
L.~Peternel, T.~Noda, T.~Petri{\v{c}}, A.~Ude, J.~Morimoto, and J.~Babi{\v{c}}, ``Adaptive control of exoskeleton robots for periodic assistive behaviours based on emg feedback minimisation,'' \emph{PloS one}, vol.~11, no.~2, p. e0148942, 2016.

\bibitem{kober2013reinforcement}
J.~Kober, J.~A. Bagnell, and J.~Peters, ``Reinforcement learning in robotics: A survey,'' \emph{The International Journal of Robotics Research}, vol.~32, no.~11, pp. 1238--1274, 2013.

\bibitem{schaal1999imitation}
S.~Schaal, ``Is imitation learning the route to humanoid robots?'' \emph{Trends in cognitive sciences}, vol.~3, no.~6, pp. 233--242, 1999.

\bibitem{nah2024modularEDA}
M.~C. Nah, J.~Lachner, F.~Tessari, and N.~Hogan, ``On the modularity of elementary dynamic actions,'' in \emph{2024 IEEE/RSJ International Conference on Intelligent Robots and Systems (IROS)}.\hskip 1em plus 0.5em minus 0.4em\relax IEEE, 2024.

\bibitem{saveriano2019merging}
M.~Saveriano, F.~Franzel, and D.~Lee, ``Merging position and orientation motion primitives,'' in \emph{2019 International Conference on Robotics and Automation (ICRA)}.\hskip 1em plus 0.5em minus 0.4em\relax IEEE, 2019, pp. 7041--7047.

\bibitem{koutras2020dynamic}
L.~Koutras and Z.~Doulgeri, ``Dynamic movement primitives for moving goals with temporal scaling adaptation,'' in \emph{2020 IEEE International Conference on Robotics and Automation (ICRA)}.\hskip 1em plus 0.5em minus 0.4em\relax IEEE, 2020, pp. 144--150.

\end{thebibliography}

\addtolength{\textheight}{-12cm}   

\end{document}